\documentclass[a4paper, 10pt, conference]{ieeeconf}      
\usepackage{FG2024}
\usepackage{float}
\usepackage{tabularray,varwidth}
\usepackage{array}
\usepackage{subfigure}
\usepackage{graphicx}
\usepackage{algorithm}
\usepackage{algorithmic}
\usepackage{amsmath}
\usepackage{amssymb}
\usepackage[switch]{lineno}
\usepackage{xcolor}

\usepackage{siunitx}
\usepackage{amsmath}

\FGfinalcopy 

\IEEEoverridecommandlockouts                              
\def\FGPaperID{121} 

\title{\LARGE \bf
One-Stage Open-Vocabulary Temporal Action Detection\\ Leveraging Temporal Multi-scale and Action Label Features
}

\author{\parbox{16cm}{\centering
    {\large Trung Thanh Nguyen$^{1, 2}$, Yasutomo Kawanishi$^{2,1}$, Takahiro Komamizu$^{3, 1}$ and Ichiro Ide$^{1, 3}$}\\
    {\normalsize
    $^1$Graduate School of Informatics, Nagoya University, Nagoya, Aichi 464-8601, Japan\\
    $^2$Guardian Robot Project, Information R\&D and Strategy Headquarters, RIKEN, Seika, Kyoto 619-0288, Japan\\
    $^3$Mathematical and Data Science Center, Nagoya University, Nagoya, Aichi 464-8601, Japan
    }}
}

\usepackage{fancyhdr}
\begin{document}

\maketitle
\thispagestyle{fancy}

\begin{abstract}
Open-vocabulary Temporal Action Detection (Open-vocab TAD) is an advanced video analysis approach that expands Closed-vocabulary Temporal Action Detection (Closed-vocab TAD) capabilities. Closed-vocab TAD is typically confined to localizing and classifying actions based on a predefined set of categories. 
In contrast, Open-vocab TAD goes further and is not limited to these predefined categories.
This is particularly useful in real-world scenarios where the variety of actions in videos can be vast and not always predictable.
The prevalent methods in Open-vocab TAD typically employ a 2-stage approach, which involves generating action proposals and then identifying those actions. 
However, errors made during the first stage can adversely affect the subsequent action identification accuracy.
Additionally, existing studies face challenges in handling actions of different durations owing to the use of fixed temporal processing methods. 
Therefore, we propose a 1-stage approach consisting of two primary modules: Multi-scale Video Analysis (MVA) and Video-Text Alignment (VTA). 
The MVA module captures actions at varying temporal resolutions, overcoming the challenge of detecting actions with diverse durations. 
The VTA module leverages the synergy between visual and textual modalities to precisely align video segments with corresponding action labels, a critical step for accurate action identification in Open-vocab scenarios.
Evaluations on widely recognized datasets THUMOS14 and ActivityNet-1.3, showed that the proposed method achieved superior results compared to the other methods in both Open-vocab and Closed-vocab settings. This serves as a strong demonstration of the effectiveness of the proposed method in the TAD task.

\end{abstract}

\section{INTRODUCTION}
Humans' ability to recognize unseen objects or actions with just their name or simple explanations stems from their capacity to apply accumulated relevant knowledge from past experiences.
In recent years, advancements in large-scale Vision-and-Language (V\&L) models have realized this capability on computers.
These models learn shared representations among images and texts by leveraging diverse image-text pairs through Contrastive Learning techniques \cite{le2020contrastive}.
As a result, these models can effectively extract valuable information from textual descriptions and visual representations for various tasks such as multimodal understanding \cite{dai2022enabling, lu2022unified, salin2022vision}, semantic comprehension \cite{ji2023seeing, li2021semvlp, li2020oscar}, few-shot learning \cite{alayrac2022flamingo, mu2019shaping, najdenkoska2023meta}, and zero-shot learning \cite{dorbala2022clip, liang2023mo, xu2023challenges}.
This capability is attained without extensive training, allowing them to exhibit excellent performance across these tasks.
Consequently, this progress has inspired researchers to explore the utilization of V\&L models in various domains, including object detection \cite{gu2021open, zareian2021open}, action recognition \cite{qian2022multimodal, weng2023transforming}, and temporal action detection \cite{ju2022prompting, rathod2022open}.

\begin{figure}
    \centering
    \subfigure[Illustration of the 2-stage approach.]
    {\includegraphics[width=0.45\textwidth]{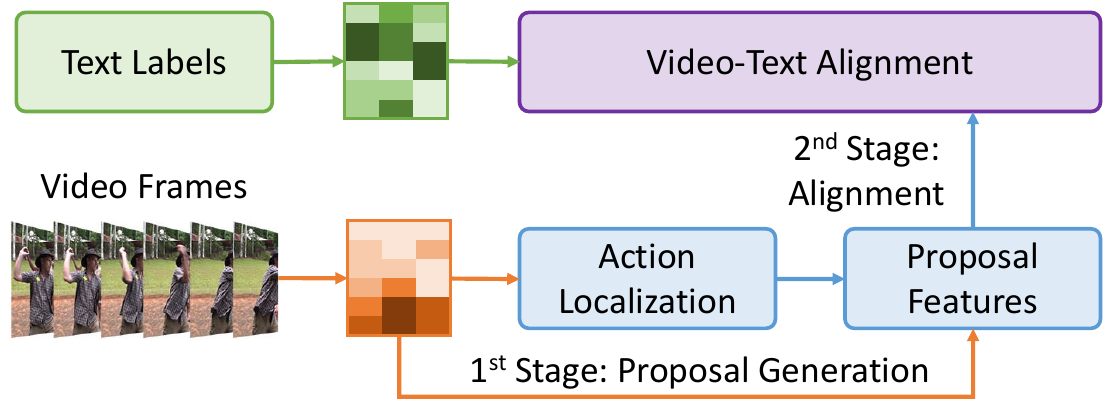}
    \label{fig:existing_approach}}
    \hfill
    \subfigure[Illustration of the 1-stage approach (Proposed method).]{
    \includegraphics[width=0.45\textwidth]{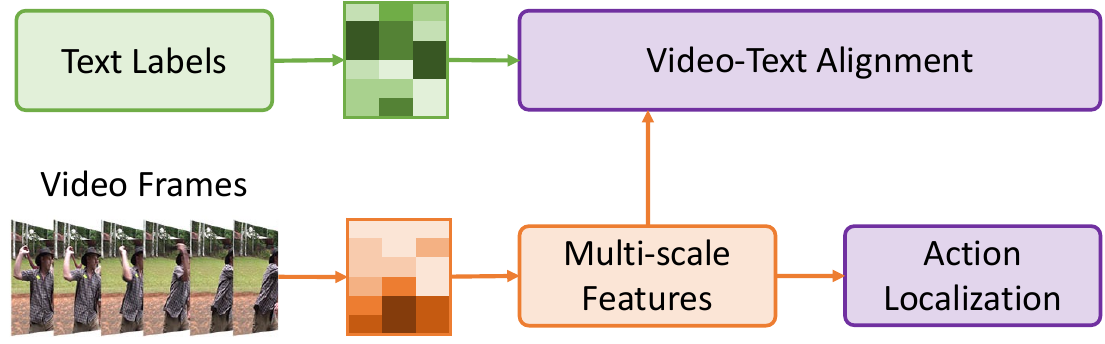}
    \label{fig:proposal_approach}}
    \caption{
    Comparing the 2-stage approach with the proposed method. (a) The 2-stage approach involves localizing temporal actions through proposal generation and utilizing the identified intervals for action identification in the alignment stage. (b) The proposed method leverages multi-scale features for both video-text alignment and action localization.}
    \label{fig:compare_method}
\end{figure}

The Closed-vocabulary Temporal Action Detection (Closed-vocab TAD) task entails localizing temporal actions within videos and classifying the corresponding action classes, assuming the action classes are defined in advance.
In contrast, Open-vocabulary Temporal Action Detection (Open-vocab TAD) requires localizing and identifying temporal actions absent from the training set.
This means that Open-vocab TAD aims to accurately detect actions that have not been previously encountered.
This poses a more challenging task, requiring the model to handle novel and unanticipated actions during the localization and identification processes.
In existing studies  \cite{ju2022prompting, rathod2022open}, a 2-stage TAD approach is adopted for Open-vocab TAD. 
In the first stage, the temporal actions are localized, and then, in the second stage, the identified interval of each action is utilized for action identification.
However, errors in the first stage can influence the accuracy of the action identification.
Moreover, existing Open-vocab TAD approaches face challenges in handling actions of different durations owing to the use of fixed temporal processing methods. 
This limitation arises when confronted with unseen actions, as the fixed temporal processing approach may not adequately capture the temporal characteristics of these novel actions. 

Figure~\ref{fig:compare_method} illustrates the comparison between the conventional and proposed methods. To address the limitations of the current Open-vocab TAD approach, we propose a 1-stage approach that combines temporal action localization and identification for the Open-vocab TAD task. 
Moreover, the proposed method aims to overcome the challenges of different durations of actions by offering a solution that implements a multi-scale component.

The contributions of this study are summarized as follows:
\begin{itemize}
    \item We propose a 1-stage approach for Open-vocab TAD that consists of two main modules: Multi-scale Video Analysis (MVA) and Video-Text Alignment (VTA), which are effective in addressing the challenges associated with Open-vocab scenarios and enabling accurate detection of a wide range of actions.
    \item We introduce a novel fusion strategy that combines temporal multi-scale features extracted from videos with action label features.
    This integration enhances the performance of action detection by effectively capturing actions of various lengths, thereby improving the accuracy and robustness of the model.
    \item We conduct extensive evaluations on widely used TAD datasets, including the THUMOS14 \cite{idrees2017thumos} and ActivityNet-1.3 \cite{caba2015activitynet} datasets. 
    Through these evaluations, we demonstrate the effectiveness of the proposed MVA and VTA modules in achieving superior performance in both Open-vocab and Closed-vocab settings.
\end{itemize}

This paper is organized as follows: 
First, we briefly summarize the relevant literature in Section~\ref{sec:related_work}.
Details of the proposed method are presented in Section~\ref{sec:proposed_method}, followed by an evaluation in Section~\ref{sec:experiments}.
Finally, Section~\ref{sec:conclusion} concludes the paper and discusses future directions.

\section{RELATED WORK}
\label{sec:related_work}
\subsection{Closed-vocabulary Temporal Action Detection (Closed-vocab TAD)}
Closed-vocab TAD focuses on action detection from untrimmed videos. 
Its approaches can be broadly classified into two types: 1-stage approaches \cite{liu2020progressive, liu2022end, zhang2022actionformer}, which are trained end-to-end and directly predict and classify action segments, and 2-stage approaches \cite{qing2021temporal, xu2020g, zhao2021video}, which employ a range of techniques to predict candidate segments and subsequently identify them using action classifiers.
In the context of the 1-stage method, a hierarchical architecture is constructed using a combination of Convolutional Neural Networks (CNNs) and Graph Neural Networks (GNNs). 
On the other hand, in the 2-stage approach, most previous studies emphasize the proposal generation phase, involving the prediction of action boundary probabilities and dense matching of start and end instants based on prediction scores.
However, these approaches rely on a predefined set of actions for both the training and inference stages, requiring careful consideration of the completeness of the action annotations.

\subsection{Transformer-based Closed-vocab TAD} 
In recent years, there has been a notable trend in Closed-vocab TAD to harness the power of transformers, driven by the remarkable success of transformers in various domains like machine translation.
Several recent studies \cite{liu2022end, shi2022react, wang2021temporal} have embraced the attention mechanism inherent in transformers to enhance the performance of action detection.
Specifically, DEtection TRansformer (DETR) \cite{carion2020end} introduces a Transformer-based approach for image detection, where it learns shared decoder input features for all input videos and detects a fixed number of outputs. 
Building upon this, Liu et al. \cite{liu2022end} proposes an end-to-end framework for Closed-vocab TAD. 
This training paradigm is known for its high efficiency and rapid prediction capabilities.
Furthermore, Zhang et al.~\cite{zhang2022actionformer} employ a transformer-based encoder to extract video representations.
In our work, we also utilize the capabilities of a Transformer-based encoder to extract video features, incorporating multi-scale features into our approach.

\subsection{V\&L models}
In recent years, there has been growing interest in V\&L models.
They can learn unified representations for both images and texts, enabling the successful completion of diverse tasks that were previously considered challenging.
Notable V\&L models include Contrastive Language-Image Pre-training (CLIP)~\cite{radford2021learning}, A Large-scale ImaGe and Noisy-text embedding (ALIGN)~\cite{jia2021scaling}, and Batch, dAta and model SIze Combined scaling (BASIC)~\cite{pham2021combined}.
The shared embedding space learned by these models from large-scale Internet datasets has enabled highly accurate Open-vocab classification tasks without fine-tuning and expensive training processes.
In the domain of videos, V\&L models have been utilized for action classification tasks \cite{ju2022prompting, wang2021actionclip} and video-text retrieval tasks~\cite{luo2022clip4clip}. 
By combining V\&L models, these approaches can recognize new actions in unseen videos containing novel scenes. This advancement paves the way for recognizing unseen actions in real-world scenarios.

\subsection{V\&L for TAD} 
Recent research in the field of V\&L applied to TAD has aimed to tackle two significant challenges: Open-set Temporal Action Detection (Open-set TAD) \cite{baoopental, chen2023cascade} and Open-vocab TAD \cite{ju2022prompting, rathod2022open}. 
Bao et al. \cite{baoopental} and Chen et al. \cite{chen2023cascade} have proposed a framework for Open-set TAD by introducing an ``Unknown'' class to handle missing classes, whereas Open-vocab TAD focuses on dealing with an open-ended vocabulary of action classes.
Both Ju et al. \cite{ju2022prompting} and Rathod et al. \cite{rathod2022open} focus on a general Open-vocab TAD and employ a 2-stage model that replaces the supervised classifier in V\&L feature comparisons. 
Our work differs from those in \cite{ju2022prompting, rathod2022open} in several aspects.
First, to avoid error propagation, a 1-stage model is adopted for both action localization and identification. 
Furthermore, to adapt to the diverse lengths of each action, multi-scale features along the temporal axis are utilized to improve the performance of Open-vocab TAD.

\section{PROPOSED METHOD}
\label{sec:proposed_method}
\begin{figure*}
    \centering
    \subfigure[Overview of the proposed method.]
    {\includegraphics[width=0.75\textwidth]{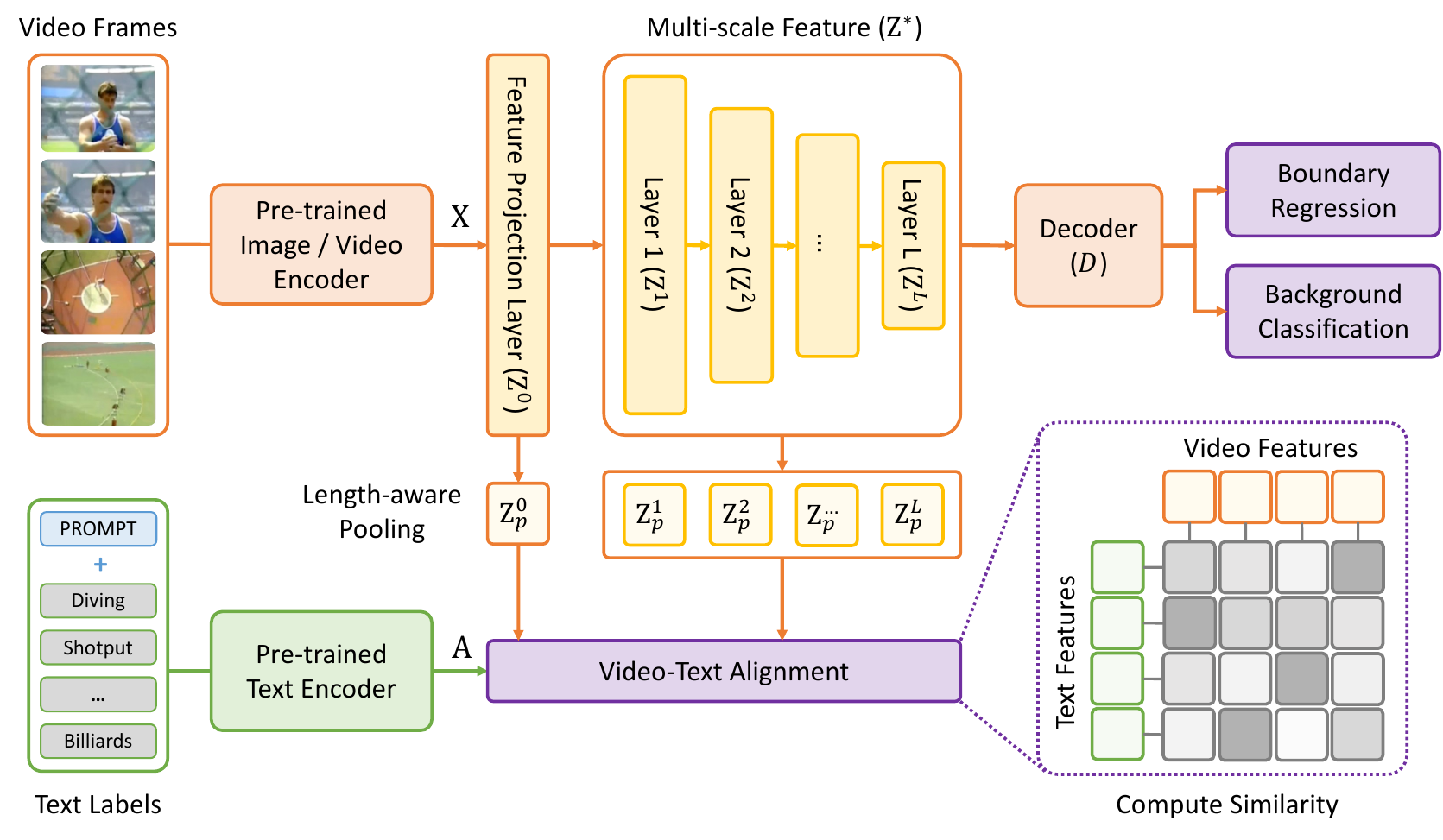}
    \label{fig:proposedmodel}}
    \hspace{1cm}
    \subfigure[Each layer in Multi-scale component.]{
    \includegraphics[width=0.16\textwidth]{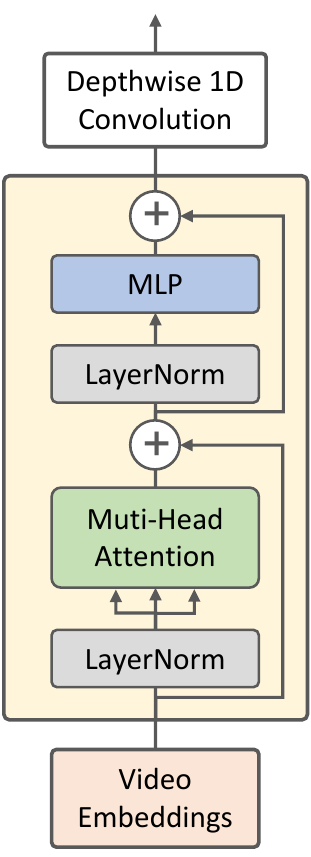}
    \label{fig:transformerencoder}}
    \caption{(a) Overview of the proposed method. Each video frame is passed through a Pre-trained Image/Video Encoder, followed by Multi-scale Video Analysis and a Decoder for segment detection. Text labels are embedded using a Pre-trained Text Encoder and aligned with the video features to comprehend the relationship and accurately determine the action labels. (b) Each layer in the Multi-scale component utilizes a Transformer Encoder for feature extraction, followed by depthwise 1D convolution for downsampling between layers.}  
    \label{fig:overview}
\end{figure*}

For given input video frames $I = (I_1, I_2, \dots, I_T)$, our objective is to estimate a series of temporal actions and corresponding classes $Y = \{\mathrm{y}_1, \mathrm{y}_2, \dots, \mathrm{y}_N\}$ within the context of Open-vocab TAD ($D_{\textrm{train}}\cap D_{\textrm{test}} = \emptyset$), which signifies that the actions present in the test set ($D_{\textrm{test}}$) do not appear in the training set ($D_{\textrm{train}}$). 
Here, $T$ is the length of the video, and $N$ is the number of temporal actions in $I$. 
We extract a feature vector $\mathbf{X} = (\mathbf{x}_1, \mathbf{x}_2, \dots, \mathbf{x}_T)$ from the input video frames $I$, where the maximum value $T$ of frames varies depending on the length of the video.
The input text labels associated with actions are also extracted as features when combined with a prompt $\mathbf{A} = \{\mathbf{a}_1, \mathbf{a}_2, \dots, \mathbf{a}_M\}$, where $M$ is the number of temporal actions in the training set ($D_\text{train}$).
These text features are integrated with video features to predict the labels of temporal actions.
The output of the model is represented as a set of $\mathrm{\textbf{y}_\textit{i}} = (s_i, e_i, a_i)$, where $s_i$ and $e_i$ denote the start and end times, respectively, and $a_i\in D_\text{train} \cup D_\textrm{test}$ is one of the action labels used for training or testing.

To elucidate the proposed method, we begin with an overview in Section~\ref{subsec:overview}, followed by detailed explanations of its critical components in Sections~\ref{subsec:mva} and \ref{subsec:vta}. Finally, we elaborate on learning objectives in Section~\ref{subsec:learningobjectives}, which play crucial roles in enhancing the overall performance.

\subsection{Overview}
\label{subsec:overview}

Figure \ref{fig:overview} illustrates an overview of the proposed method. 
We adopt a 1-stage detection approach for temporal action detection in Open-vocab TAD. The proposed method consists of two key components: (1) Multi-scale Video Analysis (MVA) and (2) Video-Text Alignment (VTA) modules. 
The former determines the start and end times of actions, as well as determines whether that time frame contains actions or not.
On the other hand, the latter is responsible for learning the relationship between video and text features to determine the labels for the actions.

\subsection{Multi-scale Video Analysis (MVA) Module}
\label{subsec:mva}

The input for the MVA module is the video frames $I$, which are then encoded through a pre-trained image/video encoder to obtain feature sequences $\mathbf{X} = (\mathbf{x}_1, \mathbf{x}_2, \dots, \mathbf{x}_T) \in \mathbb{R}^{T \times D}$. 
The encoded information is then used to construct a multi-scale feature representation $\mathbf{Z}^* = \{\mathbf{Z}^1, \mathbf{Z}^2, \dots, \mathbf{Z}^L\}$, where $L$ represents the number of scales corresponding to the hierarchical levels of the network. 
Finally, the feature representation is decoded to determine the start and end times of actions and assess the presence or absence of the action.

\subsubsection{Projection layer}
At the beginning, a shallow neural network is employed as a projection function $E: \mathbb{R}^{D} \rightarrow \mathbb{R}^{D'}$ to embed each input feature $\mathbf{x}_t$ into a $D'$-dimensional space, resulting in $\mathbf{Z}^0 = (E(\mathbf{x}_1), E(\mathbf{x}_2), \dots, E(\mathbf{x}_T))$ as the output.

\subsubsection{Multi-scale layer}
The feature vector $\mathbf{Z}^0$ serves as the input for the multi-scale component. The embedded features $\mathbf{Z}^{l}$ are then transformed into the feature representation $\mathbf{Z}^{l+1}$ of the next scale layer using the Transformer Encoder \cite{vaswani2017attention} function $f^{l+1}_{l}$ as:
\begin{equation*}
\mathbf{Z}^{l+1} = f^{l+1}_{l}(\mathbf{Z}^{l}).
\end{equation*}
It is important to note that any function, such as a CNN \cite{lea2017temporal}, GNN \cite{zeng2019graph}, can be used for this transformation. 
Figure~\ref{fig:transformerencoder} shows the architecture of the Transformer Encoder. 
The idea behind using the Transformer Encoder for the Open-vocab TAD task is to leverage the self-attention mechanism, which enables the calculation of frame similarity and the creation of a weight matrix for video frames. 
This allows automatic extraction of important frames, focusing on relevant action-containing segments and eliminating irrelevant noise. 
Within the Transformer Encoder, the Multi-head Attention mechanism captures diverse temporal dependencies between frames, while the MLP (Multi-Layer Perceptron) further refines the attended features, extracting discriminative information for precise action detection within video sequences.
In this study, we utilize a strided depthwise 1D convolution after the Transformer Encoder for downsampling between layers.
This convolution operation reduces the feature size by half in each subsequent layer. 
This process is applied for a total of $L$ layers, resulting in a multi-scale feature representation denoted as $\mathbf{Z}^* = \{\mathbf{Z}^1, \mathbf{Z}^2, \dots, \mathbf{Z}^L\}$.

\subsubsection{Decoder}
The decoder $D$ predicts sequence labeling for each frame using the multi-scale feature $\mathbf{Z}^*$.
It estimates the probabilities of action occurrence and their corresponding time intervals, denoted as $\widehat{Y} = \{\widehat{y}_1, \widehat{y}_2,\dots,\widehat{y}_N\}$.
For each $\widehat{y} = \{ d^s_t, d^e_t, p(a_t) \}$, where $d^s_t = s_t + t$ represents the time difference from frame $t$ to start time $s_t$, $d^e_t = e_t - t$ represents the time difference from frame $t$ to end time $e_t$, and $p(a_t)$  indicates the probability of the presence of an action.
The decoder incorporates a lightweight convolutional network with two heads: the \textit{Boundary Regression head} and \textit{Background Classification head}.
The former estimates each action's start and end times within the video frames $(d^s_t, d^e_t)$, while the latter predicts the probabilities of action occurrence for each frame $p(a_t)$.

\subsection{Video-Text Alignment (VTA) Module}
\label{subsec:vta}

The VTA module receives text labels of actions as input and integrates them with a prompt. These texts are transformed into text features using a pre-trained text encoder and are denoted as $\mathbf{A} = \{\mathbf{a}_1, \mathbf{a}_2, \dots, \mathbf{a}_M\}$, where $M$ corresponds to the number of classes in the training set $(D_\text{train})$.
Subsequently, the text features $\mathbf{A}$ are combined with the length-aware pooled video features $\mathbf{Z}^0_p$ and $\mathbf{Z}^l_p \in \mathbf{Z}^*_p$ to align the text and video representations.

\subsubsection{Length-aware pooling}
In the proposed method, we take action-related features to establish alignment with textual information. 
To achieve this, action representations are extracted from the video features in projection layer $\mathbf{Z}^0$ and each layer $\mathbf{Z}^l \in \mathbf{Z}^*$ based on ground-truth segments. 
Following that, average pooling is performed on the features extracted from each action interval. 
This pooling process combines the features within the intervals and generates representative features denoted as $\mathbf{Z}^0_p$ and each layer $\mathbf{Z}^l_p \in \mathbf{Z}^*_p$.
These representative features are then utilized to align with the corresponding text, facilitating the synchronization between video and text information.

\subsubsection{Video-text alignment}
The text features $\mathbf{A}$ are aligned with the video features extracted from $\mathbf{Z}^0_p$ and corresponding scale layer $\mathbf{Z}^l_p \in \mathbf{Z}^*_p$. 
We calculate the similarity between these features using the dot product, represented as $\mathbf{Z}^0_p \cdot \mathbf{A}^\top$ and $\mathbf{Z}^l_p \cdot \mathbf{A}^\top$, where $\top$ denotes the transpose operation.
The dot product measures the similarity based on the magnitude and direction of their components, indicating the level of alignment between the text and video features.
By applying the dot product, we can quantify the similarity and establish meaningful associations between textual and visual information in video-text alignment.

\subsection{Learning Objectives}
\label{subsec:learningobjectives}
\subsubsection{Objective function for MVA}
This module utilizes two loss functions to the \textit{Boundary Regression head} and \textit{Background Classification head} for every frame  $t$ $(1 \leq t \leq T)$.
The former employs Distance Intersection over Union (DIoU) loss \cite{zheng2020distance} to accurately regress the distances to the boundaries of the actions, aiming for precise localization of the actions as: 
\begin{equation}
    L^{t}_{\mathrm{BR}} = 1 - \textrm{IoU} + \mathcal{R}_{\textrm{DIoU}},
\label{eqn:loss_br}
\end{equation}
where $\textrm{IoU}$ represents the Intersection over Union between the predicted and ground-truth action interval. 
The term $\mathcal{R}_{\textrm{DIoU}}$ measures the inconsistency between the predicted interval and the ground-truth interval using the DIoU metric.
The latter employs Focal loss \cite{lin2017focal} to effectively handle imbalanced samples between background and action as:
\begin{equation}
    L^{t}_{\mathrm{BC}} = -\alpha_t(1-p_t)^\gamma\log(p_t),
\label{eqn:loss_bc}
\end{equation}
where $p_t$ is the predicted probability of an action occurrence, $\alpha_t$ is the balancing factor to address the class imbalance, and $\gamma$ is a modulating factor that focuses on hard samples.
The overall loss function of the MVA module ($L_{\textrm{MVA}}$) is defined as the sum of the above losses (Eqn.~\ref{eqn:loss_br} and Eqn.~\ref{eqn:loss_bc}) for each time step as:
\begin{equation}
L_{\textrm{MVA}} = \sum_{t}^{T}{(L^t_{\mathrm{BR}} + \lambda_1 L^t_{\mathrm{BC}})},
\label{eq:mva}
\end{equation}
where $\lambda_1$ is a balancing coefficient.

\subsubsection{Objective function for VTA}
This module aims to model the cross-modal relationship between two modalities: video features and text features. 
The principle is to minimize the distance between representations of corresponding video-text pairs, while encouraging those of non-corresponding pairs.
The learning objective $L_\text{VTA}$ consists of two contrastive terms: $L_{\mathbf{Z}^0 \to \mathbf{A}}$ (Eqn.~\ref{eq:z_to_a}) and $L_{\mathbf{Z}^* \to \mathbf{A}}$ (Eqn.~\ref{eq:z*_to_a}). The former aligns the video projection features with text features, while the latter aligns the video multi-scale features with text features. Below are the details of the VTA loss:
\begin{equation}
L_{\mathbf{Z}^0 \to \mathbf{A}} = - \frac{1}{N} \sum_{i=1}^{N} \log \frac{\exp((\mathbf{Z}^0_p \cdot \mathbf{A}^\top)^+ / \tau)}{\sum_{j=1}^{M} \exp((\mathbf{Z}^0_p \cdot \mathbf{A}^\top)^- / \tau)},
\label{eq:z_to_a}
\end{equation}
\begin{equation}
L_{\mathbf{Z}^* \to \mathbf{A}} = - \frac{1}{N} \sum_{i=1}^{N} \sum_{l=1}^{L} \log \frac{\exp((\mathbf{Z}^l_p \cdot \mathbf{A}^\top)^+ / \tau)}{\sum_{j=1}^{M} \exp((\mathbf{Z}^l_p \cdot \mathbf{A}^\top)^- / \tau)},
\label{eq:z*_to_a}
\end{equation}
\begin{equation}
L_{\text{VTA}} = L_{\mathbf{Z} \to \mathbf{A}} + \lambda_2 L_{\mathbf{Z}^* \to \mathbf{A}},
\label{eq:vta}
\end{equation}
where $N$ is the number of temporal actions, $M$ is the number of text classes, $L$ is the number of multi-scale layers, ``$+$'' and ``$-$'' represent a pair of samples that are corresponding and non-corresponding, respectively, $\tau$ is a temperature hyperparameter controlling the impact of penalties on hard negative samples, and $\lambda_2$ is a balancing coefficient.

\subsubsection{Overall objective function}
The overall loss function of the proposed method is defined as the sum of the MVA loss (Eqn.~\ref{eq:mva}) and VTA loss (Eqn.~\ref{eq:vta}), multiplied by a balance coefficient $\lambda_3$ as:
\begin{equation}
    L_\text{Total} = L_\text{MVA} + \lambda_3 L_\text{VTA}.
\label{eq:total}
\end{equation}

\begin{table*}[th]
\caption{Results on THUMOS14 and ActivityNet-1.3 datasets. mAP $(\uparrow)$ at different tIoU thresholds are reported. The average mAP in the range of [0.30:0.10:0.70] is reported for THUMOS14, while [0.50:0.05:0.95] is reported for ActivityNet-1.3. Best results within each group are highlighted in \textbf{bold}, while the overall best results are \underline{underlined}.}
\subtable[Results in a completely Open-vocab setting using text features from pre-trained CLIP base and pre-trained CLIP large.]
{
\begin{tblr}{
  colspec = {|Q[l,1.9cm]|Q[c,1.38cm]|Q[c,1.38cm]|Q[r,0.55cm]Q[r,0.55cm]Q[r,0.55cm]Q[r,0.55cm]|Q[r,0.55cm]Q[r,0.55cm]Q[r,0.55cm]Q[r,0.55cm]|Q[c,0.8cm]|Q[c,0.8cm]|Q[c,0.8cm]|},
}
\hline
\SetCell[r=3]{l}{Model} & \SetCell[r=3]{c}{Image \\ Feature} & \SetCell[r=3]{c}{Text \\ Feature} & \SetCell[c=8]{c}{THUMOS14 \cite{idrees2017thumos}} & & & & & & & & \SetCell[c=3]{c}{ActivityNet-1.3 \cite{caba2015activitynet}} & \\ \hline
 &  &  & \SetCell[c=4]{c}{75/25} & & & & \SetCell[c=4]{c}{50/50} & & & & Smart & 75/25 & 50/50 \\ \hline
& & & 0.3 & 0.5 & 0.7 & Avg. & 0.3 & 0.5 & 0.7 & Avg. & Avg. & Avg. & Avg. \\
\hline
\SetCell[r=4]{l}{OV-TAD \cite{rathod2022open}} & CLIP B/16  & CLIP B/16 & 21.8 & 13.2 & 3.7 & 12.9 & 18.0 & 8.9 & 2.2 & 9.5 & 23.4 & 21.4 & 19.5\\ 
 & I3D  & CLIP B/16 & 27.8 & 15.3 & 4.3 & 15.6 & \textbf{23.6} & 12.8 & 3.2 & 12.9 & 24.1 & 22.1 & 20.1 \\ 
 & CLIP B/32  & CLIP B/32 & 21.4 & 11.8 & 3.5 & 12.0 & 15.4 & 7.7 & 1.9 & 8.0 & 22.6 & 19.4 & 17.3  \\ 
  & I3D  & CLIP B/32 & 25.1 & 13.9 & 4.0 & 14.1 & 21.0 & 11.3 & 3.0 & 11.5 & 23.9 & 20.2 & 18.2 \\ \cline{1}
Baseline & I3D & CLIP B/16 & 24.3 & 16.6 & 5.6 & 15.8 & 15.5 & 10.7 & 3.8 & 10.2 & 21.7 & 16.7 & 14.5  \\
Proposed & I3D& CLIP B/16 & \textbf{32.1} & \textbf{25.3} & \textbf{\underline{13.6}} & \textbf{24.0} & 18.6 & \textbf{15.2} & \textbf{9.3} & \textbf{14.6} & \textbf{28.2} & \textbf{22.6} & \textbf{20.8} \\ 
\hline
 \SetCell[r=2]{l}{OV-TAD \cite{rathod2022open}} & CLIP L/14  & CLIP L/14 & 28.6 & 15.4 & 4.2 & 15.8 & 21.0 & 9.8 & 2.0 & 10.5 & 28.7 & 24.6 & 22.4\\ 
 & I3D  & CLIP L/14 & 30.1 & 16.8 & 4.7 & 17.0 & \textbf{\underline{26.1}} & 14.3 & 3.6 & 14.5 & 28.1 & 24.8 & \textbf{\underline{22.8}}\\ \cline{1}
 Baseline & I3D & CLIP L/14 & 29.5 & 19.7 & 5.6 & 18.8 & 18.3 & 11.2 & 3.1 & 10.9 & 21.0 & 17.4 & 14.4 \\ 
Proposed & I3D & CLIP L/14 & \textbf{\underline{35.3}} & \textbf{\underline{26.6}} & \textbf{\underline{13.6}} & \textbf{\underline{25.5}} & 18.7 & \textbf{\underline{15.8}} & \textbf{\underline{9.6}} & \textbf{\underline{15.0}} & \textbf{\underline{28.8}} & \textbf{\underline{24.9}} & 21.1 \\
\hline
\end{tblr}
\label{tab:result_main_1}
}
\hfill
\subtable[Results of methods that incorporate the classification score with the weakly-supervised action recognition model \cite{wang2017untrimmednets}.]{
\begin{tblr}{
  colspec = {|Q[l,1.9cm]|Q[c,1.38cm]|Q[c,1.38cm]|Q[r,0.55cm]Q[r,0.55cm]Q[r,0.55cm]Q[r,0.55cm]|Q[r,0.55cm]Q[r,0.55cm]Q[r,0.55cm]Q[r,0.55cm]|Q[c,0.8cm]|Q[c,0.8cm]|Q[c,0.8cm]|},
}
\hline
EffPrompt~\cite{ju2022prompting} & I3D & CLIP B/16 & 39.7 & 23.0 & 7.5 & 23.3 & 37.2 & 21.6 & 7.2 & 21.9 & --- & 23.1 & 19.6 \\
STALE~\cite{nag2022zero} & I3D & CLIP B/16 & 40.5 & 23.5 & 7.6 & 23.8 & 38.3 & 21.2 & 7.0 & 22.2 & --- & 24.9 & 20.5  \\
Baseline & I3D & CLIP B/16 & 51.6 & 29.7 & 7.1 & 29.5 & 44.2 & 24.5 & 5.7 & 24.8 & 25.3 & 23.0 & 21.9 \\
Proposed & I3D & CLIP B/16 & \textbf{60.4} & \textbf{47.4} & \textbf{24.3} & \textbf{44.8} & \textbf{51.9} & \textbf{38.1} & \textbf{17.9} & \textbf{36.5} & \textbf{36.0} & \textbf{34.1} & \textbf{33.6} \\
\hline
\end{tblr}
\label{tab:result_main_2}
}
\label{tab:result_main}
\end{table*}

\section{EVALUATION}
\label{sec:experiments}
In this section, we evaluate the proposed method extensively through a series of comprehensive experiments. In addition, we perform in-depth ablation studies to gain clearer insights into the key characteristics of the proposed method.

\subsection{Experimental Conditions}

\subsubsection{Datasets}
We perform evaluations on the THUMOS14 \cite{idrees2017thumos} and ActivityNet-1.3 \cite{caba2015activitynet} datasets, which are widely used for TAD.
\begin{itemize}
    \item THUMOS14 dataset contains 413 videos with 20 action categories, with an average of 15 instances per video.
    \item ActivityNet-1.3 dataset is a large-scale action dataset, consisting of 200 activity classes and approximately 20,000 videos with more than 600 hours.
\end{itemize}

\subsubsection{Data preparation}
We decompose these datasets into train-test splits by following the Open-vocab data split strategy \cite{rathod2022open}. 
This approach randomly splits action categories into specific ratios, with corresponding videos associated with these splits. 
According to \cite{rathod2022open}, we choose two ratios, namely, ``75/25 split'' and ``50/50 split''.
The former involves 10 random splits, selecting 75\% of action categories and corresponding videos as a training set and the rest as a test set. 
Similarly, the latter involves 5 random splits, selecting 50\% for a training set and the rest as a test set.
In addition, with the Acitivity-Net1.3 dataset, we employ the ``Smart split'' strategy described in \cite{rathod2022open}, which leverages the hierarchy of action categories, with 25\% of the labels allocated to the test set. The smart split is constructed by selecting pairs of neighboring leaf nodes from the hierarchy, designating one node for evaluation, and including the other in the training set. The selection process considers the perceptual similarities between classes to ensure an effective split.

\subsubsection{Evaluation metrics}
To evaluate the final output of the model, we utilize Soft Non-Maximum Suppression (Soft-NMS) \cite{bodla2017soft} to eliminate duplicate detections, considering the potential overlap among the estimated action candidates. 
Subsequently, the results are evaluated based on the mean Average Precision (mAP), which measures the percentage of correctly estimated actions using threshold processing on the temporal Intersection over Union (tIoU) with ground truth. 

\subsubsection{Comparison methods}
We compare the proposed method with the following methods:
\begin{itemize}
    \item Baseline model: An 1-stage approach similar to the proposed model but utilizes a single scale within the multi-scale component to observe the effectiveness of the multi-scale component of the proposed method.
    \item OV-TAD \cite{rathod2022open}: The current state-of-the-art Open-vocab TAD setting, operating in a 2-stage TAD approach.
    \item STALE \cite{nag2022zero} and EffPrompt \cite{ju2022prompting}: 1-stage and 2-stage Open-vocab TAD methods that incorporate the classification score, respectively, with the weakly-supervised action recognition model  \cite{wang2017untrimmednets}. We also utilize these scores for comparison when comparing with these methods.
\end{itemize}

\subsubsection{Implementation details}
The proposed method was implemented according to the detailed description in Section~\ref{sec:proposed_method}, utilizing six layers in the MVA module. 
For the video encoder, we utilize a two-stream Inflated 3D (I3D) ConvNet pre-trained on the Kinetics dataset~\cite{carreira2017quo} to extract video features. 
For the text encoder, we utilize a pre-trained CLIP model \cite{radford2021learning} to extract text features. 
Following Rathod et al. ~\cite{rathod2022open}, we do not use a prompt in the main experiment. The text prompt is analyzed in Section~\ref{subsubsection:text_prompt}.
In the projection layers, we use fully connected layers with Gaussian Error Linear Unit (GELU) activation \cite{hendrycks2016gaussian}.
The balance coefficients in Eqns. (\ref{eq:mva}), (\ref{eq:vta}), and (\ref{eq:total}) are set as 1. We investigate the impact of the coefficients in Section~\ref{subsubsection:change_lambda}. 
For optimization purposes, we utilize AdamW \cite{loshchilov2017decoupled}. 
To attain optimal results, we carefully considered each dataset's model complexity and available training data, ensuring the appropriate selection of hyperparameters.

\subsection{Experimental Results}
Table~\ref{tab:result_main} presents the performance of the proposed method with other comparison methods. 
We report the mAP at different tIoU thresholds, with the average calculated in the range of mAP@[0.30:0.10:0.70] for THUMOS14 and mAP@[0.50:0.05:0.95] for ActivityNet-1.3 datasets. 
The results are divided into two tables; Table~\ref{tab:result_main_1} shows the results in a completely Open-vocab setting, while Table~\ref{tab:result_main_2} shows the results with methods using fusion classification scores from a weakly-supervised action recognition model.
In Table~\ref{tab:result_main_1}, the results are split into two groups using pre-trained text features from CLIP base and CLIP large models.

\subsubsection{THUMOS14}
The proposed method consistently attained the highest results for both groups shown in Table~\ref{tab:result_main_1}, surpassing other methods by a considerable margin.
The results indicated a significant advantage of the proposed method, particularly at mAP@0.7, where it outperformed other methods nearly threefold. 
This demonstrated the substantial contribution of the multi-scale feature in achieving more accurate unseen action detection.
In Table~\ref{tab:result_main_2}, since the classes of actions in a THUMOS14 dataset video are mostly singular, the average mAP of the proposed method using fusion classification scores was more than twice that without using them. 
Furthermore, the proposed method's performance was nearly $1.5$ to $2.0$ times higher compared to the STALE \cite{nag2022zero} and EffPrompt \cite{ju2022prompting} methods.

\subsubsection{ActivityNet-1.3}
In both groups in Table~\ref{tab:result_main_1}, the proposed method consistently outperformed other methods in most of the split scenarios.
Particularly in the Smart split, the proposed method demonstrated performance gains ranging from 4.1\% to 6.5\% in the pre-trained CLIP base group and from 0.1\% to 7.8\% in the pre-trained CLIP large group.
Furthermore, the proposed method also demonstrated notable growth in the 75/25 split and 50/50 split. 
In contrast, the baseline model exhibited suboptimal performance on this dataset, which can be attributed to the dataset's large size and diverse nature, including a significant number of labels that were not sufficiently trained during the training process. 
These results underscore the significant contribution of the multi-scale feature in enabling the proposed method to overcome these challenges and achieve superior results.
In Table~\ref{tab:result_main_2}, the proposed method was also superior to other benchmarks when considering the fusion classification score.

\subsection{Ablation Study}
We conducted a series of excision experiments as part of an ablation study to evaluate the effectiveness of the proposed method on the THUMOS14 dataset using the ``75/25 split'' and ``50/50 split'' evaluation settings.

\begin{table}[tb]
\centering
\caption{Effect of Changes of feature extraction in MVA module and fusion strategy in VTA module. Average mAP $(\uparrow)$ in the range of [0.30:0.10:0.70] is reported.} 
\begin{tblr}{
  colspec = {|l|c|c|c|c|},
}
\hline
\SetCell[r=2]{c}{} & \SetCell[c=2]{c}{CLIP B/16} & & \SetCell[c=2]{c}{CLIP L/14}  \\ \hline
 & 75/25 & 50/50 & 75/25 & 50/50 \\ \hline
 Proposed & \textbf{24.0} & \textbf{14.6} & \textbf{25.5} & \textbf{15.0} \\
 Trans Layer $\to$ Conv Layer & 22.4  & 12.8 & 23.2 & 12.9 \\
 (w/o) Projection Feature ($\mathbf{Z}^0_p$) & 21.7  & 12.3 & 23.0 & 14.4 \\
 (w/o) Multi-scale Feature ($\mathbf{Z}^*_p$) & 22.5 & 12.2 & 24.7 & 14.0 \\
\hline
\end{tblr}
\label{tab:fusion_strategy_backbone}
\end{table}

\subsubsection{Feature extraction in the MVA module and fusion strategy in the VTA module}
Table~\ref{tab:fusion_strategy_backbone} shows the effects of modifying components within the MVA and VTA modules. Specifically, when we replaced Transformer layers (Trans Layer) with Convolutional layers (Conv Layer), we observed that the Trans Layer outperformed the Conv Layer, highlighting the beneficial impact of the attention mechanism. Additionally, we investigated altering the fusion strategy between text and video features. Utilizing either the individual projection feature ($\mathbf{Z}^0_p$) or the multi-scale feature ($\mathbf{Z}^*_p$) independently resulted in reduced performance compared to integrating both with the text feature. These results underscore the importance of harnessing the combined strength of temporal multi-scale features and action label features to enhance performance significantly.

\begin{table}[tb]
\centering
\caption{Effect of Number of layers in the multi-scale feature of the MVA module. Average mAP $(\uparrow)$ in the range of [0.30:0.10:0.70] and inference time are reported.} 
\begin{tblr}{
  colspec = {|c|c|c|c|c|c|},
}
\hline
\SetCell[r=2]{c}{Number \\of layers} & \SetCell[c=2]{c}{CLIP B/16} & & \SetCell[c=2]{c}{CLIP L/14} & & \SetCell[r=2]{c}{Inference Time [s] \\ (B/16 $\sim$ L/14)} \\ \hline
& 75/25 & 50/50 & 75/25 & 50/50 \\ \hline
1 & 17.5 & 10.5 & 18.3 & 10.0 & \textbf{0.1038 $\sim$  0.1202}  \\
3 & 21.0 & 11.4 & 23.4 & 11.8 & 0.1280 $\sim$ 0.1428 \\
5 & 23.5 & 13.6 & 24.8 & \textbf{15.1} &  0.1774 $\sim$ 0.1813 \\
6 & 24.0 & \textbf{14.6} & 25.5 & 15.0 & 0.1786 $\sim$ 0.1878 \\ 
7 & \textbf{24.3} & 14.2 & \textbf{26.8} & 14.8 & 0.2013 $\sim$ 0.2458 \\\hline
\end{tblr}
\label{tab:number_layer}
\end{table}


\subsubsection{Number of layers in the MVA module}
Table~\ref{tab:number_layer} shows the performance impact of varying the number of layers in the MVA module.
The results indicate that increasing the number of layers consistently improved performance, as evidenced by the increase in mAP.
However, the model tended to converge in terms of mAP at layers 5, 6, and 7, where the performances were relatively similar.
Adding more layers beyond a certain point does not yield significant performance improvements.
There was a slight trade-off with the inference time as the computational complexity increased with more layers.  
Specifically, seven layers required more than twice the computation of one layer.
Therefore, selecting an optimal number of layers in the MVA module is crucial to balance performance and computational efficiency.

\begin{table}[tb]
\centering
\caption{Impact of text prompts. Average mAP $(\uparrow)$ in the range of [0.30:0.10:0.70] is reported.} 
\begin{tblr}{
  colspec = {|l|c|c|c|c|},
}
\hline
\SetCell[r=2]{l}{Prompt} & \SetCell[c=2]{c}{CLIP B/16} & & \SetCell[c=2]{c}{CLIP L/14} & \\ \hline
 & 75/25 & 50/50 & 75/25 & 50/50 \\ \hline
$[\textsc{Class}]$ & 24.0 & 14.6 & 25.5 & 15.0 \\
 A video of action $[\textsc{Class}]$ & 24.3 & 12.8 & 26.8 & 13.5 \\
 $[\textsc{Class}] + [\textsc{Description}]$ & \textbf{28.7} & \textbf{16.1} & \textbf{29.6} & \textbf{16.2} \\
 $[\textsc{Description}]$ & 25.0 & 13.3 & 27.2 & 14.5 \\
\hline
\end{tblr}
\label{tab:prompt}
\end{table}

\subsubsection{Text prompts}
\label{subsubsection:text_prompt}
We evaluated the impact of text prompts on the Open-vocab TAD accuracy. 
The results, as shown in Table IV, indicate that a basic prompt such as ``A video of action $[\textsc{Class}]$'' yields negligible improvements. 
The THUMOS14 dataset contains simplistic labels like Shot Put, High Jump, and Long Jump, which lack detailed contextual information. The experiments with $[\textsc{Class}] + [\textsc{Description}]$ prompts, with descriptions generated from ChatGPT \cite{OpenAIChatGPT3.5} like ``\textit{High Jump: Athlete jumps over a horizontal bar}'', significantly enhanced the detection performance by approximately 4\% in the 75/25 split and 1.5\% in the 50/50 split. 
However, the results also underscore that merely adding descriptions without the action class is insufficient to substantially increase the accuracy of action detection, pointing to the necessity of a nuanced approach in employing text prompts for video analysis.

\begin{figure}[t]
    \includegraphics[width=0.5\textwidth]{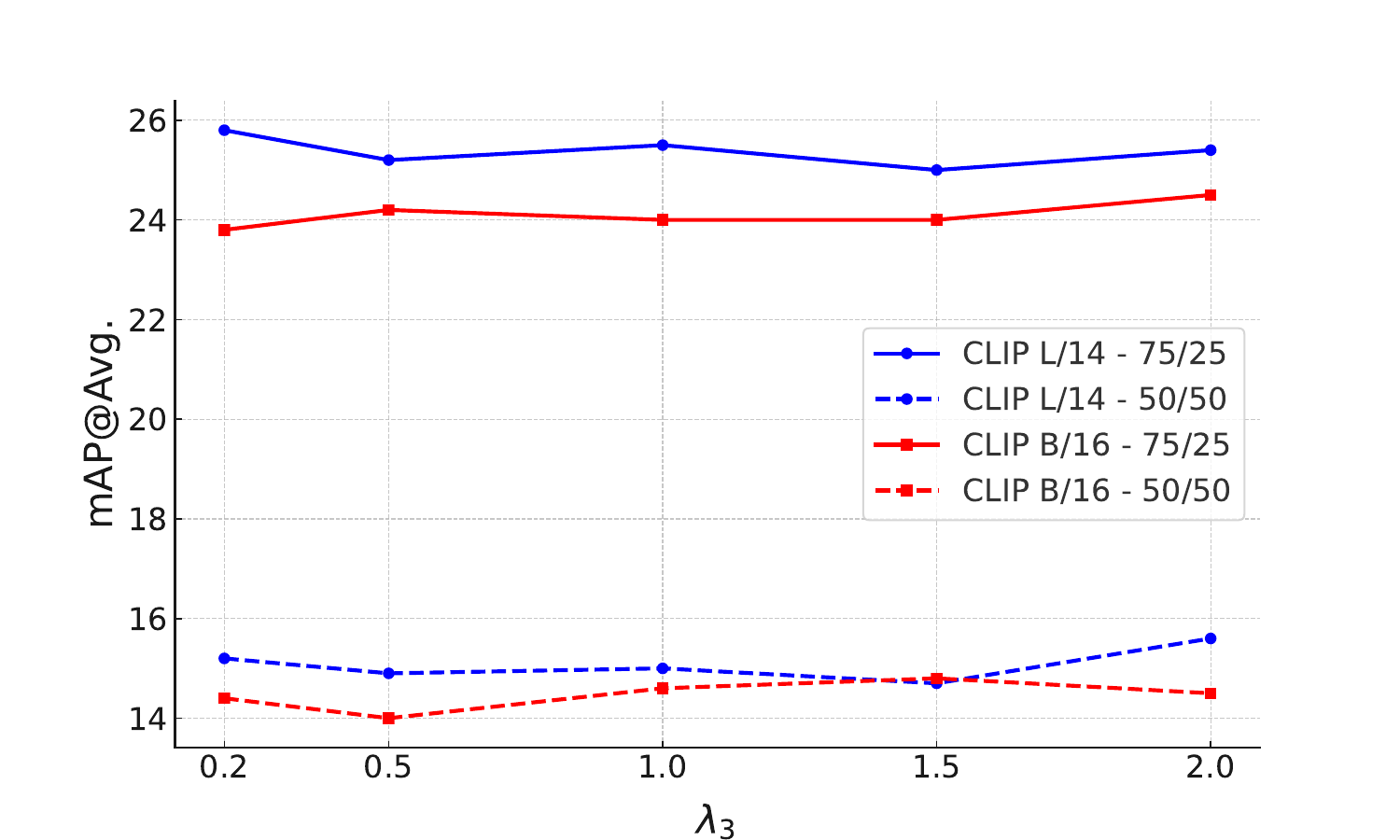}
    \caption{Change of $\lambda$ coefficient in the overall loss function.}  
    \label{fig:change-lambda}
\end{figure}
\subsubsection{$\lambda$ coefficient in the overall loss function}
\label{subsubsection:change_lambda}
We experimented on the interplay between MVA Loss and VTA Loss as formulated in Eqn.~\ref{eq:total}. The outcomes of this analysis are visually represented in Fig.~\ref{fig:change-lambda}. The results demonstrate that varying the value of $\lambda_3$ from $0.2 \to 2.0$ only resulted in a deviation of no more than $\pm 0.5$ compared to that when $\lambda_3$ was set to 1. This indicates a relatively stable performance of the loss function across a range of $\lambda$ coefficient values.

\begin{figure*}
    \centering
    \subfigure[Illustrating action detection of the ``Shot Put'' event.]{
        \includegraphics[width=0.45\textwidth]{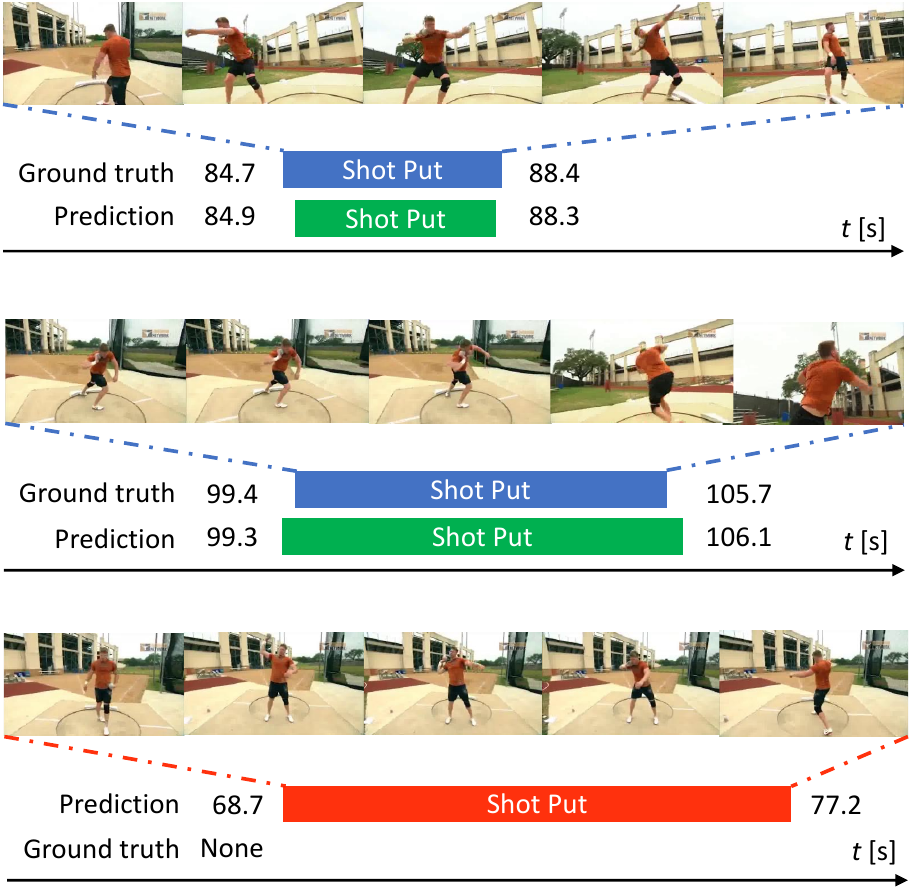}
        \label{fig:visualizaiton_shot_put}
    } 
    \subfigure[Illustrating action detection of the ``High Jump'' event.]{
        \includegraphics[width=0.45\textwidth]{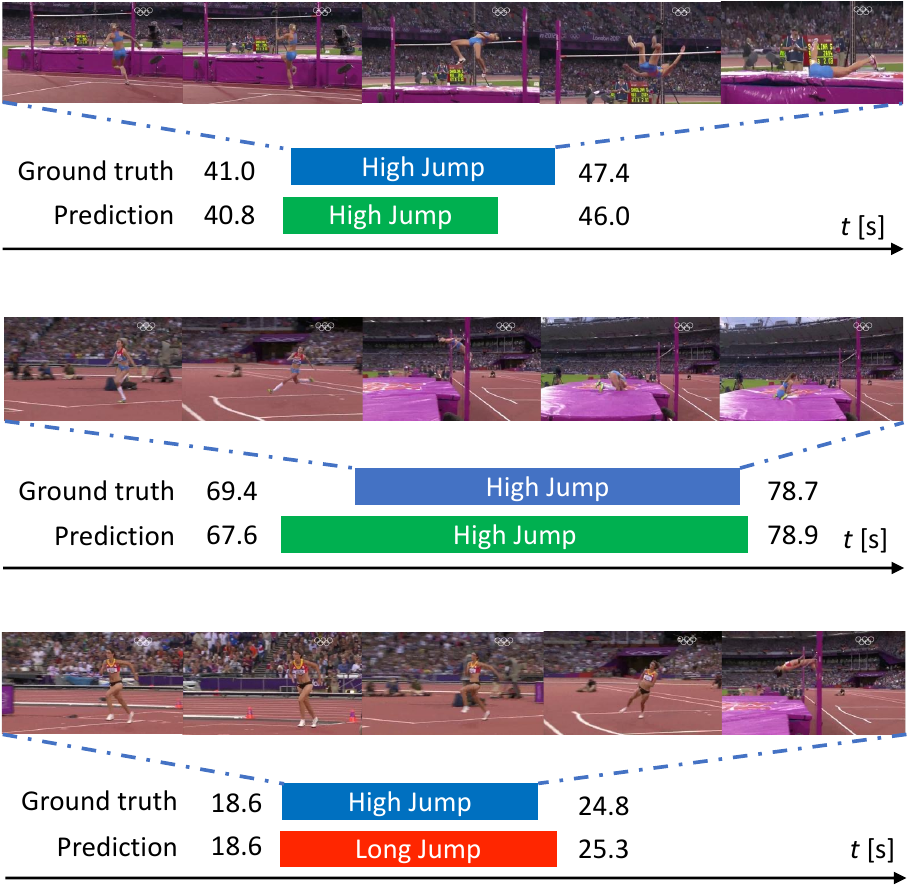}
        \label{fig:visualization_high_jump}
    }
    \caption{Illustrating the results of Open-vocab TAD using CLIP B/16 as the text feature.}  
    \label{fig:visuzaliation}
\end{figure*}

\subsection{Open-vocab TAD Results Visualization}
We present the visualization of outcomes for several actions in Fig.~\ref{fig:visuzaliation}. In Fig.~\ref{fig:visualizaiton_shot_put}, the results for the ``Shot Put'' action were identified almost accurately across varying action durations. However, the third action scene showed an error in action localization where the video frames predominantly displayed the athlete performing preparatory motions similar to the ``Shot Put'' action, leading to a wrong detection. 
Similarly, in Fig.~\ref{fig:visualization_high_jump}, the ``High Jump'' action was also detected nearly accurately in the first two scenes, with varying lengths of the action. 
The third scene wrongly identified where it should have been ``High Jump'' was as ``Long Jump''. 
This error occured because most of the frames in this sequence showed the athlete running, which was a characteristic more associated with ``Long Jump'', leading to the misidentification of the action. 
These visualizations highlight the effectiveness and limitations of Open-vocab TAD in accurately localizing and identifying actions within a given video, underscoring the importance of nuanced differentiation between similar actions and the challenges posed by actions with similar preparatory movements.

\subsection{Closed-vocab Temporal Action Detection Setting}
\subsubsection{Setting}
In this section, we evaluate the Closed-vocab setting ($D_{\textrm{test}}\subset D_{\textrm{train}}$), which refers to the common context in which the model undergoes training and evaluation using the same action categories. 
It is important to note that the common context utilizes only videos and lacks the ability to detect unseen actions. 
In contrast, the Open-vocab setting employs both video and text, enabling evaluation of seen and unseen actions.
To ensure a fair and consistent comparison, we use the same dataset splits as those utilized in previous studies and evaluate them on the THUMOS14 dataset.

\subsubsection{Comparison methods}
We considered the following methods for conducting comparisons with the proposed method. 
In the common context, some of the modern models for TAD in recent years have utilized the Inflated 3D (I3D) ConvNet and Temporal Segment Network (TSN) encoder backbone.
In the Open-vocab setting, we conducted comparisons with methods using various pre-trained text-image models. 
The baseline model remains unchanged, as previously mentioned.

\begin{table}[tb]
\centering
\caption{Comparison in a Closed-vocab setting using the THUMOS14 dataset. 
Average mAP $(\uparrow)$ in the range of [0.30:0.10:0.70] is reported.} 
\begin{tblr}{
  colspec = {|l|c|c|c|c|},
}
\hline
\SetCell[r=2]{l}{Model} & \SetCell[r=2]{c}{Setting} & \SetCell[r=2]{c}{Image\\ Feature} & \SetCell[r=2]{c}{Text\\ Feature} & \SetCell[r=2]{c}{mAP\\@Avg.}  \\ \hline
& & & & \\ \hline
BMN \cite{lin2019bmn} & \SetCell[r=7]{c}{Video} & TSN & --- & 38.5 \\ 
TAL-MR \cite{zhao2020bottom} & & I3D & --- & 43.3 \\ 
VSGN \cite{zhao2021video} & & TSN & --- & 50.2 \\ 
AFSD \cite{lin2021learning} &  & I3D & --- & 52.0 \\ 
TadTR \cite{liu2022end} & & I3D & --- & 46.6 \\ 
ActionFormer \cite{zhang2022actionformer} & & I3D & --- & 66.8 \\
TriDet \cite{shi2023tridet} & & I3D & --- & \textbf{69.3} \\ \hline
EffPrompt \cite{ju2022prompting} & \SetCell[r=8]{c}{Video \\ \& \\ Text}  & CLIP B/16 & CLIP B/16 & 34.5 \\ 
STALE \cite{nag2022zero} & & CLIP B/16 & CLIP B/16 & 44.4 \\
STALE \cite{nag2022zero} & & I3D & CLIP B/16 & 52.9 \\ 
OV-TAD \cite{rathod2022open} & &  CLIP B/32 & CLIP B/32 & 26.6 \\ 
OV-TAD \cite{rathod2022open}  & & CLIP B/16 & CLIP B/16 & 29.0 \\ 
OV-TAD \cite{rathod2022open}  & & CLIP L/14 & CLIP L/14 & 32.6 \\ 
Baseline & & I3D & CLIP B/16 & 39.9 \\
Proposed & & I3D & CLIP B/16 & \textbf{59.5} \\ \hline
\end{tblr}
\label{tab:closed-set}
\end{table}

\subsubsection{Results}
The results presented in Table~\ref{tab:closed-set} show that the proposed method achieved the highest mAP@Avg of 59.5\% in the video \& text setting (Open-vocab methods). 
This performance is particularly notable compared to leading methods using only video settings (Closed-vocab methods) such as ``ActionFormer'' and ``TriDet''. 
The results of the proposed method were closely competitive, and it outperformed approximately two-thirds of the existing methods. 
Meanwhile, the baseline model only achieved 39.9\%, underscoring the significance of the 19.6\% performance gap due to its lack of multi-scale component. 
These results highlight that the proposed method excels among the Open-vocab methods and performs well among the Closed-vocab methods.


\section{CONCLUSION}
\label{sec:conclusion}

We proposed a method for the Open-vocab TAD task that leveraged temporal multi-scale and action label features. 
The proposed 1-stage approach consists of a Multi-scale Video Analysis (MVA) module and a Video-Text Alignment (VTA) module.
We also introduced a fusion strategy that combined temporal multi-scale features and action label features to improve the accuracy and robustness of action detection. 
A series of comprehensive experiments on THUMOS14 \cite{idrees2017thumos} and ActivityNet-1.3 \cite{caba2015activitynet} datasets indicated that the MVA module’s multi-scale feature with the attention mechanism and the VTA module were instrumental in boosting performance. 
The number of layers in the MVA module significantly affected the experimental outcomes, necessitating a careful selection to balance performance with computational complexity. 
Furthermore, the use of text prompts within the VTA module impacted action identification results, highlighting the need for detailed analysis in Open-vocab setting.

In future work, we plan to investigate further methods for incorporating contextual information and higher-level scene understanding to enhance the performance of the action detection system. 
We also aim to address the accurate detection of actions' start and end times, as this significantly impacts the precise determination of those actions. 
Additionally, we aim to explore techniques for handling complex and dynamic scenes in cluttered or occluded environments.

\section*{Acknowledgment}
This work was partly supported by JSPS KAKENHI JP21H03519 and JP24H00733. The computation was carried out using the General Projects on supercomputer ``Flow'' at Information Technology Center, Nagoya University.

{\small
\bibliographystyle{ieee}
\bibliography{ref}
}

\end{document}